# Enhancing Machine Learning Performance with Continuous In-Session Ground Truth Scores: Pilot Study on Objective Skeletal Muscle Pain Intensity Prediction

**Boluwatife E. Faremi [1,\*], Jonathon Stavres [2], Nuno Oliveira [2], Zhaoxian Zhou [1] and Andrew H. Sung [1]**

[1] School of Computing Sciences and Computer Engineering, University of Southern Mississippi, Hattiesburg, MS 39401, USA
[2] School of Kinesiology and Nutrition, University of Southern Mississippi, Hattiesburg, MS 39401, USA
\* Correspondence: boluwatife.faremi@usm.edu

**Abstract:** Machine learning (ML) models trained on subjective self-report scores struggle to objectively classify pain accurately due to the significant variance between real-time pain experiences and recorded scores afterwards. This study developed two devices for acquisition of real-time, continuous in-session pain scores and gathering of ANS-modulated endodermal activity (EDA). The experiment recruited N = 24 subjects who underwent a post-exercise circulatory occlusion (PECO) with stretch, inducing discomfort. Subject data were stored in a custom pain platform, facilitating extraction of time-domain EDA features and in-session ground truth scores. Moreover, post-experiment visual analog scale (VAS) scores were collected from each subject. Machine learning models, namely Multi-layer Perceptron (MLP) and Random Forest (RF), were trained using corresponding objective EDA features combined with in-session scores and post-session scores, respectively. Over a 10-fold cross-validation, the macro-averaged geometric mean score revealed MLP and RF models trained with objective EDA features and in-session scores achieved superior performance (75.9% and 78.3%) compared to models trained with post-session scores (70.3% and 74.6%) respectively. This pioneering study demonstrates that using continuous in-session ground truth scores significantly enhances ML performance in pain intensity characterization, overcoming ground truth sparsity-related issues, data imbalance, and high variance. This study informs future objective-based ML pain system training.

## 1. Introduction

Pain is reported to be the center cog of opioid crisis and the majority of those suffering from opioid use disorder (OUD) state it all started from using opioids for pain with legitimate prescription [1]. In [2], authors reveal that about 9.7 million individuals aged 12 or above have misused prescribed opioids, leading to the center for disease control (CDC) declaring an opioid epidemic. Nonetheless, opioid remains the frontline treatment in the United States irrespective of several clinical guidelines recommending non-pharmacological alternatives [2, 3, 4]. In [2], authors divulge that orthopedic pain (34.8 %) is a primary reason for opioid dispensing trailed by dental pain (17.3 %), back pain (14.0 %) and headache (12.9 %). When abused, harms arising from opioid intake are but not limited to overdose, addiction, diversion, and death [1, 5]. These side effects make pain measurement clinically relevant by medical practitioners to prevent the absolute risk of death [6, 7].

Consequently, pain and its intensity are described as a personal experience that can only be described by the individual experiencing pain and its often assessed by pain intensity rating scales [8]. Typically, traditional pain rating or self-report scales such as visual analog scales (VAS), numerical rating scales (NRS) are acknowledged to be highly subjective, prone to anchor drift over time and significant variation in reported scores [6, 7, 8]. Additionally, these scales are notorious for misleading practitioners into over and under prescribing opioids, encouraging catastrophizing amongst patients, and sheltering of social-ecological construct such as race, implicit and explicit bias amongst dishonest healthcare providers [1, 9, 10]. Thereupon, objective pain assessment is suggested as an avenue to prevent drug abuse by providing decision support cues to opioid dosing to pain intensity numbers or range [11]. To achieve this, current research efforts are exploring the use of machine learning for diagnosis and management of pain [3]. These trends are infusing one or more objective measure of pain outlined in [12], one or more ground truth scores described in [13, 14] and machine learning algorithms to iteratively learn complex objective features to a class, develop predictive models that independently respond to newly unscripted objective pain data [3, 15, 16].

Despite the efficacy of previous works in Table 1, a latent bottleneck that affects the predictive performance of these objective systems was revealed and confirmed in [13, 14, 17, 18]. Authors divulged that the present variety of ground truths are sparse and only provide measurement once per stimulus or after a procedure. Consequently, leading to an unreliable measurement that causes label noise and high variation with respect to measured objective features [7, 13, 14, 17]. Validating this claim, it was announced in a study that trained machine learning models for objective pain intensity classification yielded low performance due to presence of inconsistency and variance in the periodical ground truth scores obtained from traditional or self-reporting scale (VAS) [18]. Notably, pain is not an impulse but a gradually declining response that stops after inducement, measuring pain intensity periodically after a stimulus or experiment may lead to inaccurate recall of scores due to multitasking challenges and declining response respectively [19]. As reported in [13, 17, 19], self-reporting scales do not effectively measure diverse pain intensity during procedures because of its discrete measurement, leading to class imbalances in data due to sparse information, and ostracization of important pain moments.

Hence, it is crucial in the field of objective pain assessment using machine learning to ascertain the effectiveness of combining subjective in-session continuous scores or subjective post-session scores with objective pain features. Additionally, it is important to this field to explore if the maximum of in-session continuous scores or its integral over a window length gives the best corresponding ground truth scores to objective features acquired during an experiment. As far as the author is aware this is the first study to design devices that collect both continuous in-session scores and an objective physiological marker of pain over the same time spectrum to yield better ML prediction performance.

## 2. Materials and Methods
### 2.1. Participants
This study was conducted at the physiology sensing laboratory of the Kinesiology and Nutrition Department of University of Southern Mississippi (USM), USA under an approved institutional review board protocol #22-1012. A total of twenty-four healthy volunteers with a mean age of twenty-eight were recruited to undergo Test 2 under the visit 2 section of the IRB protocol. All participants were required to declare any medical history, implanted devices, and use of any cardiovascular drugs. Moreover, all participants were provided with an informed consent, briefed about the experimental phases of the study, and introduced to all devices necessary for the experiment.

### 2.2. Experiment Design
The safe and repeatable experiment designed in this study is the post exercise circulation occlusion with muscle stretch (PECO + Stretch). In Test 2, each participant was instructed to perform a series of maximal handgrip contractions thrice with a five-minute recovery interval to obtain 35 % of the mean maximum voluntary contraction (MVC). Afterwards, each subject was given a baseline rest period for 1 minute before being instructed to perform a submaximal handgrip contraction and hold for 2 minutes (Figure 1). Just before the end of the handgrip contraction, an occlusion cuff was inflated around the upper arm for 3 minutes. During the final minute of the cuff inflation, an investigator stretched the forearm by moving the wrist, after which the cuff was deflated for a recovery period of 2 minutes. Through these protocols muscular skeletal pain was evoked through the sensitization of group III and IV afferent muscle endings respectively [20, 21].

### 2.3. Data Acquisition and Instrumentation
In [13, 14], authors implied that the development of an objective pain system necessitates the acquisition of ground truth scores alongside one or several identified objective signals identified in [12]. In this study, the PECO + Stretch experiment involved participants lying horizontally on a flat bed, while their faces were directed upward towards a screen projecting 35% MVC from LabChart Software. Subsequently, a Smedley handgrip dynamometer retrofitted with a potentiometer was fitted to the non-dominant arm of each participant to apply forces around the projected MVC setpoint. An E20 DE Hokanson rapid occlusion cuff inflator was then applied to the upper arm,

Table 1: Review of related works of machine learning for pain assessment.

| Study | Objective (Obj) | Type of Pain (TP) | Input modality — Objective Signals Used (OS) | Machine Learning Model — Type of model (TM) | Machine Learning Model — Task and way (W) | Ground truth Modality — Type of ground truth (TGT) and name (N) | Ground truth Modality — Nature of ground truth | Ground truth Modality — Mode of collection (MC) | Accuracy Result |
|---|---|---|---|---|---|---|---|---|---|
| [18] | PSISA | H | SC | SVML, SVMR, MLP, RF | 3* C, R | S, VAS | Continuous | PI | 69.7 %, 69.2 % |
| [22] | OHPA | CPS | EEG, EM, SC, BVP, EMG, RR, ST, BP, FE | GA, SVM | 3* C | S, VRS | Ordinal | PI | 73.25 %, 61.67 %, 69.83 % |
| [23] | BPC | B | KJA | MLP, RF | 2* C | S, SBST | Categorical | PS | 60 %, 70 % |
| [24] | LBPC | B | fMRI, HRV | SVM | 2* C | S, NRS | Ordinal | PS | 92.45 % |
| [25] | LBPC | B | EMG | LR | 7* C, 9* C | S, NRS | Ordinal | PS | 91.2 %, 96.7 % |
| [26] | GBPC | CP | fMRI | DL, SVM | 2* C | S, SL | Categorical | BS | 69 %, 86 % |
| [27] | HPC | HOA | EEG | SVM | 2* C | S, NRS | Ordinal | BS,PS | 79.6 % |
| [28] | SCPSP | SCP | $S_pO2$, DBP, SBP, HR, T | MLR, KNN, SVM, RF | 11* C | S, NRS | Ordinal | U | 68.1 % |
| [29] | PPC | CP | EEG | SVM | 3* C | S, VAS | Continuous | U | 71.9 % |
| [30] | PPOC | APP | MCS, EEG, PPG, SC | SVM, LR, DT, RF | 3* C, 4* C | S, VAS | Continuous | PS | 89.5 %, 76.7 % |
| [31] | PPOC | APP | SC | KNN, RF | 2* C | S, NRS | Ordinal | PS | 86 % |
| [32] | MPI | EP | BVP, ECG, SC | LDA, KNN, SVM | 4* C | U | U | U | 75 % |
| [33] | PPOC | MMP | ECG, EMG, PPG, SC | KNN, RF, SVMR, ADA | 2* C | S, NRS | Ordinal | PI | 80 % |
| [34] | PPOC | MMP | PPG | ADA, KNN, RF, SVM, XGB | 2* C | S, NRS | Ordinal | PI | 81.41 % |

**Obj**: OHPA: objective human pain assessment BPC: back pain classification LBPC: low back pain classification GBPC: Gender-based pain classification HPC: Hip pain classification SCPSP: sickle cell pain score prediction PPC: pain phenotype classification PPOC: post-operative pain classification PSISA: pain stimulation intensity and sensation assessment MPI: measurement of pain intensity **TP:** H: Heat CPS: cold pain stimulus B: Back CP: Chronic Pain HOA: hip osteoarthritis APP: acute post-operative pain EP: electrical pulse MMP: mild to moderate pain **OS**: FE: facial expression EMG: electromyography SC: skin conductance ECG: electrocardiography KJA: kinematic joint angle fMRI: functional magnetic resonance imaging HRV: Heart rate variability SBP: systolic blood pressure DBP: diastolic blood pressure HR: heart rate T: temperature $S_pO2$: oxygen saturation MCS: muscle corrugator supercilia PPG: photoplethysmogram BVP: Blood volume pressure BP: blood pressure EM: eye movement ST: skin temperature **TM:** GA: genetic algorithm RF: random forest SVM: support vector machine SVML: support vector machine linear kernel SVMR: support vector machine radial basis kernel MLP: multi-layer perceptron LR: logistic regression KNN: k-nearest neighbor DT: decision tree MLR: multinominal logistic regression LDA: linear discriminant analysis ADB: adaboost XGB: xgboost **Task:** C: classification, R: regression **W**: 2*: 2-way 3*: 3-way 4*: 4-way 7*: 7-way 9*: 9-way 11*: 11-way **TGT:** O: observer S: self-report EKP: expert and prior knowledge **N**: VRS: verbal rating scale VAS: visual analog scale SBST: STarT screening tool NRS: numerical rating scale SL: sex label **MC:** BS: before session CI: continuous in-session PI: Periodical In-session PS: post session U: Unreported and a tensile wrist cuff was placed on the non-dominant arm to establish the maximum force eliciting the slightest discomfort. For subjective signal acquisition, each participant was provided with an instrumented continuous, in-session ground truth acquisition device known as "PainSlide Meter (PSM)" (see following sub-section).

The endodermal activity (EDA) modulated by the autonomic nervous system (ANS) was selected as the chosen single objective signal [35]. Following established publication recommendations for EDA measurement, a specific area just above the abductor hallucis of each participant's inner foot was prepared with a 70% Isopropyl alcohol pad [36]. AgCl electrodes were then placed on the prepared site for EDA measurement using an exosomatic direct current (DC) sensor from [37]. The experimental phases, as depicted in Figure 1, were subsequently carried out by each participant, during which both objective and PSM in-session subjective data were acquired through an instrumented device called "PainGad" into a custom-designed pain platform named "PainPlat". Afterwards, a VAS scale was given to each participant post-experiment to record a post-session ground truth. Figure 2 depicts the entire pipeline utilized in this study.

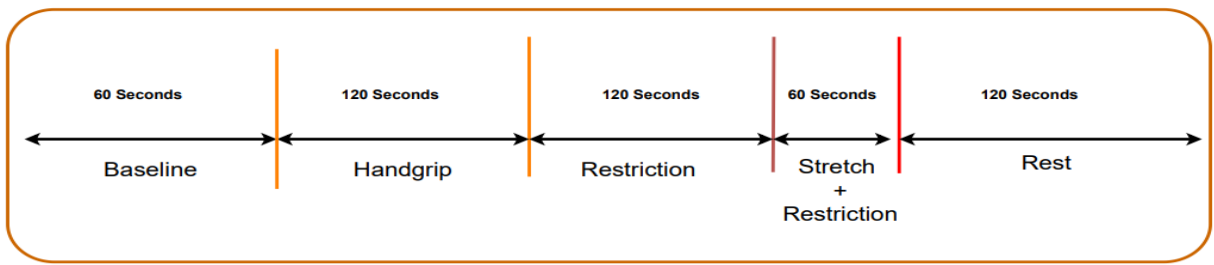

Figure 1: Experimental phases in the present study

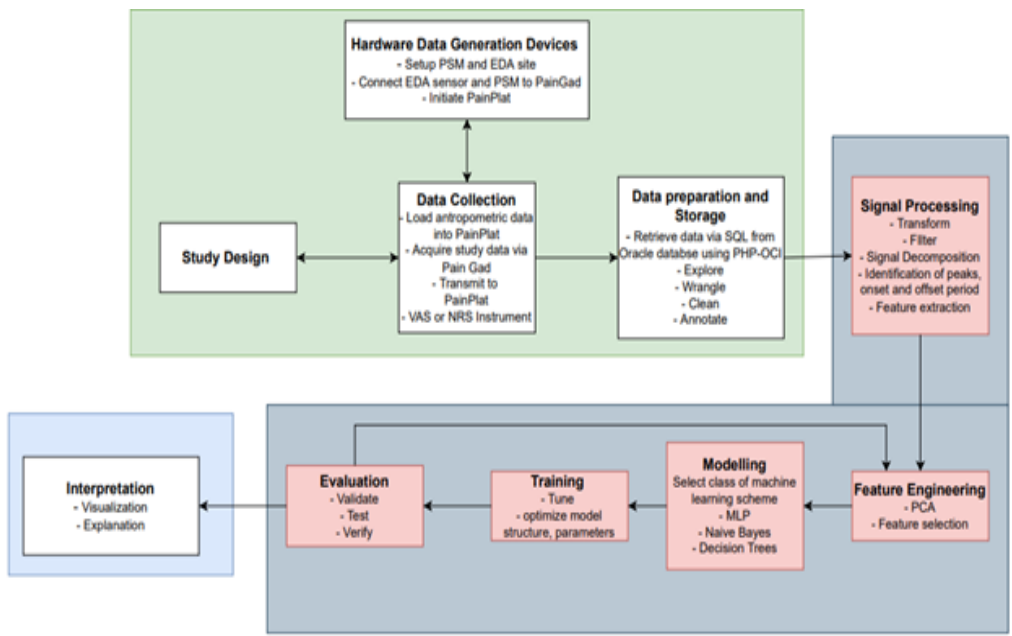

Figure 2: Pipeline utilized in the present study

### 2.3.1. Instrumentation of an In-Session Ground Truth Acquisition Device

PSM is a passive ML-centric device that continuously harvests in-session, subjective pain score over an entire time spectrum of a pain experiment and it is founded on the theory of intermodal comparison. This theory explains the mechanism through which pain intensity is mapped into a different sensory domain. In conventional settings, it can be argued that the VAS or NRS scales utilize this technique in prompting participants to map subjective pain felt in the body into a verbal, visual, or numerical quantifiable domain [38, 39]. Figure 3 depicts both the schematic diagram and instrumented PSM device. To use the PSM device, desired DC power is supplied by an active device to the intermodal sensor (linear taper slider) for digital translation of subject felt pain between the pain and no-pain continuum of the device. Additionally, a digital feedback indicator is provided to allow each participant to understand selected pain reading along the continuum. This device can be toggled between a continuous VAS mode or an ordinal NRS mode. In the present work, the former was utilized for all participants.

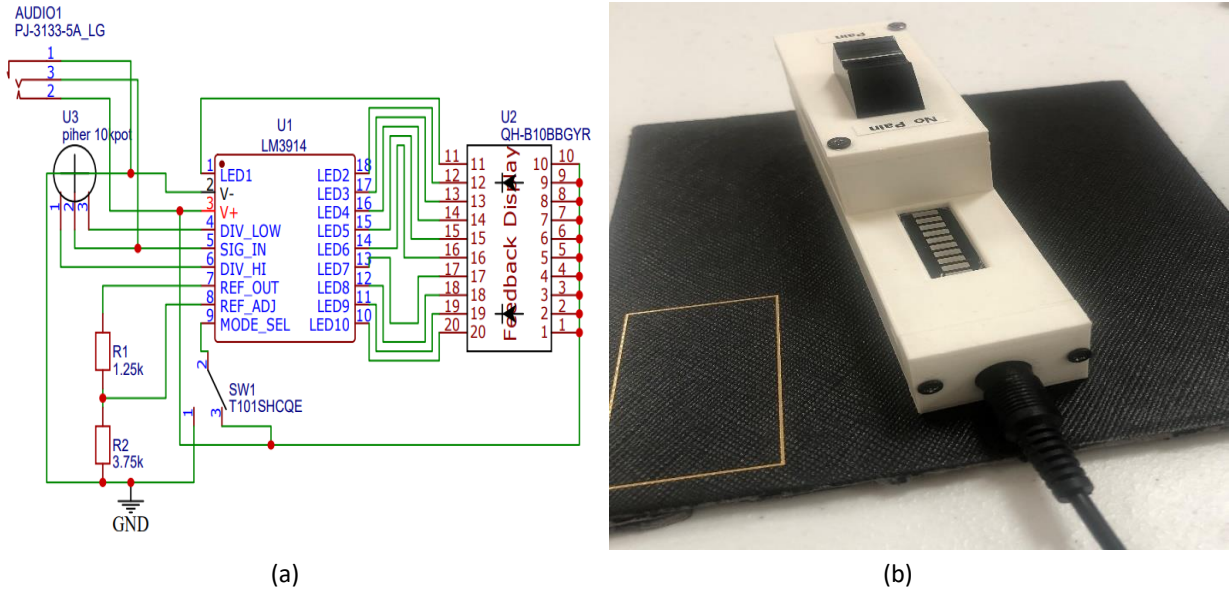

Figure 3: PSM device (a) Schematic diagram (b) Instrumented passive PSM device

As stated in literature, pain scores left in the digital form lack a form of measurement unit [38]. Consequently, mapping functions were used for the PSM device for conversion of analog voltage marked on the slider into a digital form and finally into a 0 – 10 cm scale via equations (1) – (2)

$$P_{digital} = \frac{P_{analog}}{P_{ref}} \times 2^n - 1 \qquad (1)$$

$$P_{value} = (P_{digital} - P_{gad_{min}}) \times \frac{P_{scale_{max}} - P_{scale_{min}}}{P_{gad_{max}} - P_{gad_{min}}} + P_{scale_{min}} \qquad (2)$$

where $P_{digital}$ is the digital value equivalent of subject expressed pain level on PSM, $P_{analog}$ (V) denotes the analog voltage equivalent of subject expressed pain level over PSM no pain and pain spectrum, $P_{ref}$ (V) signifies PSM direct current power supply value, $n$ represents the number of bits of the analog-to-digital (AD) channel of the pain gathering device, $P_{scale_{(.)}}$ explains the target conversion scale for a marked pain level on PSM (depending on the study this could be a 0 – 10 scale as with the VAS or scaled as 0 – 100 cm or %), $P_{scale_{max}}$ and $P_{scale_{min}}$ expresses the maximum and minimum attainable indices for $P_{scale_{(.)}}$ respectively, $P_{gad_{min}}$ gives the minimum $P_{digital}$ value that can be obtained on PSM for no pain, $P_{gad_{max}}$ indicates the maximum $P_{digital}$ value that can be obtained on PSM for high pain, and $P_{value}$ connotes final equivalent subject felt pain during the experiment on the chosen $P_{scale_{(.)}}$,

### 2.3.2. Instrumented Concomitant Data Acquisition Device and Pain Platform

A customized pain platform called PainPlat was developed as a computerized means to store quality data during the PECO + Stretch experiment for each participant [40]. The platform was implemented as a simplistic two-layer web architecture topology that combines three components of application, presentation, processing, and database in two modalities on a machine with specifications Intel dual core, 4 Gb RAM, 1.6 GHz processor speed, Windows 10 operating system, Apache 2.4.54 server and Oracle Database 18c Express Edition Production. Bootstrap 5 framework containing user interface (UI) components such as HTML, CSS, and JavaScript was used to render the application to the investigator [41, 42, 43]. Experimental parameters (such as exerted handgrip force, restriction pressure, and stretch force) that described the pain experiment for each participant were securely inputted into the platform. Considering the ethos of confidentiality and privacy, integrity constraints were used to enforce participants identifiers to 22-102-S1XXX format in the database. Thereafter, the investigator initiates the concomitant streaming device from the web-application to start sampling and storing EDA and PSM data through written oracle client instance (OCI) and PHP 8.1.13 representation state transfer (REST) application programming interfaces (API's) that was tested via Postman 10 software during development.

PainGad is a streaming device developed for concomitant acquisition of subjective in-session ground truth labels and objectively measured EDA signals from a participant. The device was designed on a dual core Espressif ESP-32 chip that comes with functionalities such as 2.4 GHz 802.11 b/g/n wifi radio, Bluetooth 4.2 BR/EDR and Bluetooth LE, 18 channels of 12-bit SAR analog to digital converter (ADC), 4 Serial Peripheral Interface, 2 I2S, 2 I2C, 3 UART [44]. In the present study, five interfaces were created in the PainGad device for the purpose of interfacing with the EDA sensor, PSM device and external data acquisition system. These interfaces are one TTRS-USB-serial communication port, two female 3.5mm TRS audio-to-BNC jack ports, one female TRS jack for the passive PSM device and a female TRRS EDA connector. At the TRRS-EDA port or TRS-PSM interface, a maximum sampling rate of 2 mega-samples per second (MSPS) can be obtained. However, owing to the webserver's limitation of handling a maximum of 150 samples per minute per HTTP thread connection, the EDA and PSM in-session scores were retrieved at a sampling frequency of 2 Hz in this study. Figure 4 depicts the Velcro strapped PainGad device interfaced with both the EDA sensor over the abductor hallucis and PSM device respectively. Figure 5 illustrates the web architecture of PainPlat receiving EDA and PSM data.

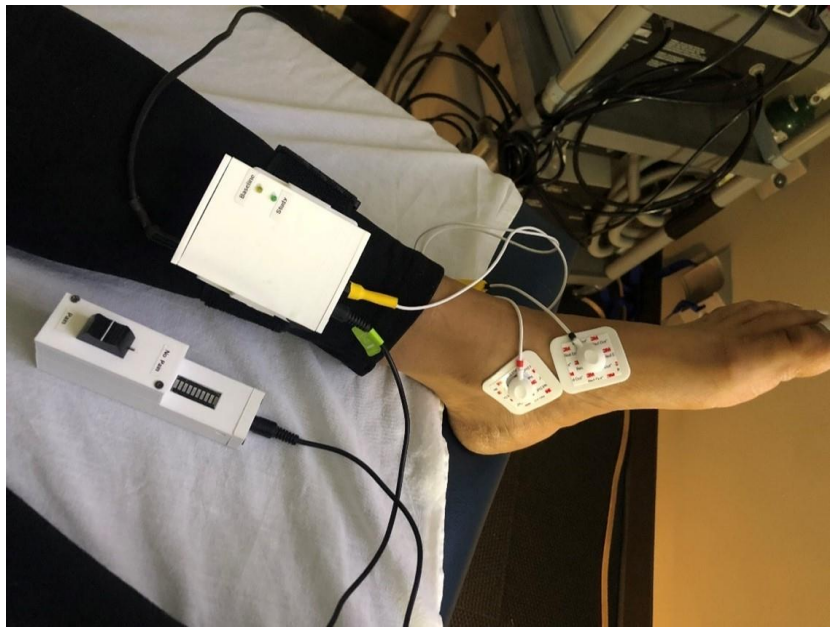

Figure 4: Instrumented PainGad device fitted around the ankle region of a participant.

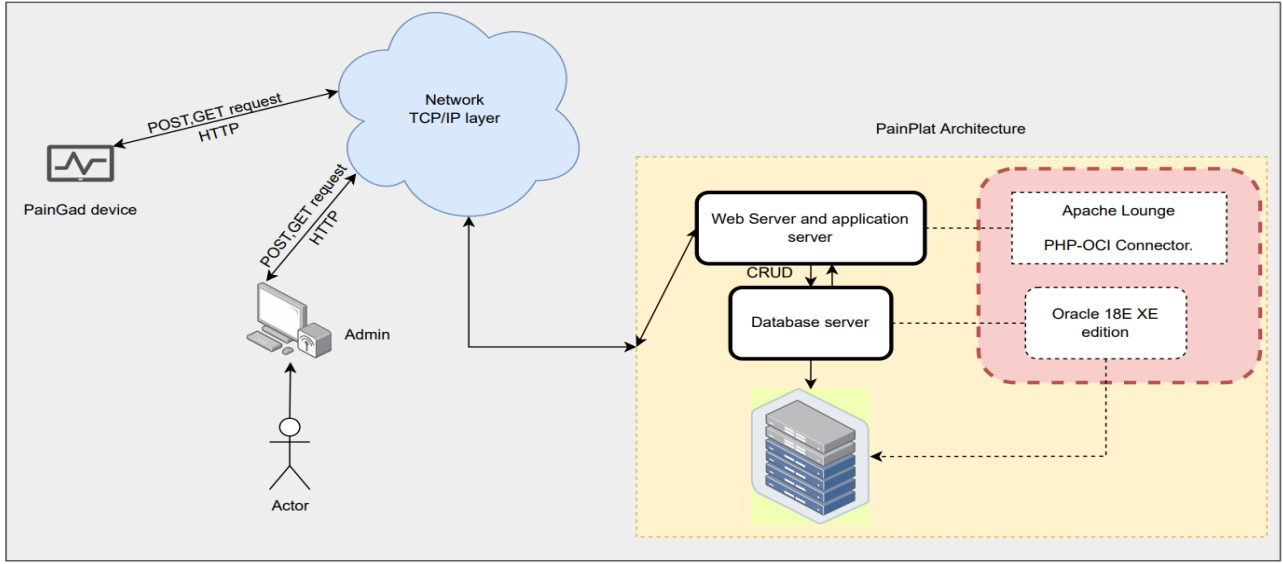

Figure 5: Developed PainPlat web-architecture for storing of experimental parameters from investigators, and streaming of PSM and EDA data from PainGad device

## 2.4. Data Processing
### 2.4.1. Data Preparation and Signal Processing.

Data stored in the PainPlat for each participant were extracted and sorted using via the structured query language (SQL). 9 participants were excluded from further processing due to history of known hypertensive drug, negligible sympathetic arousal of the eccrine gland, unstable feet, and inability to complete all phases of the experiment in Figure 1. 120 samples annotated as baseline were removed from each participant raw EDA data prior to transformation into equivalent skin conductance signals via the transfer function given in [37]. Raw $P_{digital}$ values for each participant were transformed into 0 – 10 scores using equation (2) before a median filter of interval +/- 10 second-centered on every current sample was utilized to smoothen ripples and false readings that may have occurred during baseline and experiment. Equally, 120 samples indicating baseline PSM values were removed for each participant to obtain an equal vector length for skin conductance and in-session scores before signal processing.

Decomposition of skin conductance into tonic (i.e., slowly varying) and phasic (i.e., rapidly varying) components has been widely conducted in literature [35, 45, 46]. Accordingly, custom decomposition and processing pipelines described in [46, 47, 48] were written and implemented in MATLAB to identify peaks greater than the threshold of 0.01μS in the phasic signal. Peak associated time-domain features recommended in [49, 50], were extracted over a 10 second non-overlapping window length. Alike, corresponding in-session ground truth scores for each participant were obtained during the same window length of 10 seconds as with the objective features. In each window, each participant continuous in-session scores were acquired in two folds namely PSM-Mean (integral) and PSM-Mode (maximum). Thereafter, post-session VAS ground truth scores indicated by each participant were also appended to their respective processed datasets. In sum, all participants dataset (11 features and 3 ground truths) were joined together to form a matrix size of $555 \times 14$ from original 43,200 samples. To ready the matrix for feature selection, the $555 \times 14$ matrixes were regularized using the min-max normalization before separating into 3 folds based on the ground truth types i.e., in-session ground truth scores PSM-Mean, PSM-Mode, and post-session VAS scores (See Table 2). Figure. 6 depicts skin conductance signal and corresponding unfiltered and filtered in-session ground truth scores.

Table 2: Populated Feature Vector and Ground Truth Scores

| Dataset Categories | Features | Target Ground truth |
|---|---|---|
| EDA-1 | Sum of peaks, mean, maximum peak, number of peaks, mean absolute, root mean square, minimum peak, standard deviation, force, occlusion pressure and muscle tension. | VAS Scores (i.e., scores noted by participants after experiments). |
| EDA-2 | Sum of peaks, mean, maximum peak, number of peaks, mean absolute, root mean, square, minimum peak, standard deviation, force, occlusion pressure and muscle tension. | PSM Mean Score |
| EDA-3 | Sum of peaks, mean, maximum peak, number of peaks, mean absolute, root mean, square, minimum peak, standard deviation, force, occlusion pressure and muscle tension. | PSM Mode Score |

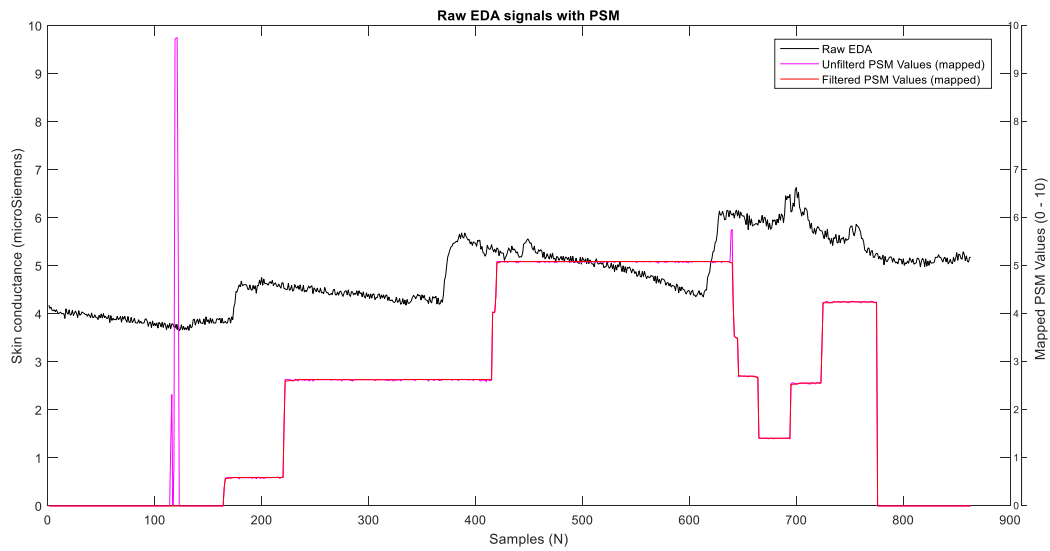

(a)

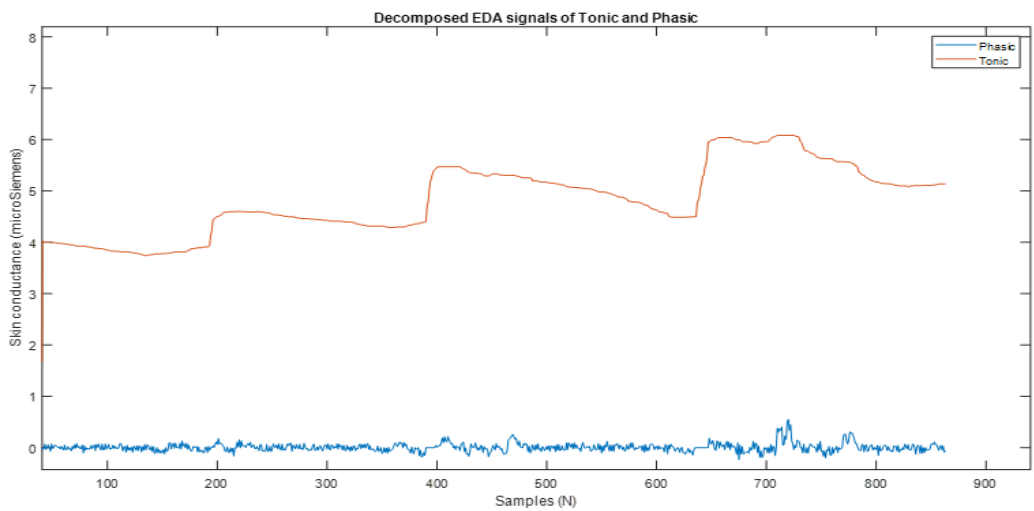

(b)

Figure 6: (a) Corresponding Trend of Skin Conductance Signal and Unfiltered/Filtered In-Session Ground Truth Scores for a Participant (b) Tonic and Phasic component of the EDA data

### 2.4.2. Feature Selection

Feature selection is a technique utilized for uncovering a subset of features that supplies meaningful information to training machine learning algorithms [51]. According to [52], performing feature selection improves accuracy of machine learning algorithms, reduces the problem of overfitting, and longer training time. Accordingly, in the present work, the filter selection method was used to determine appropriate features for training ML algorithms. The method works by ranking the p – values of all data categories with respect to a target ground truth variable [51]. Figure 7 depicts the feature ranking importance for all dataset categories. Irrelevant EDA – 1 feature (muscle tensile stretch, occlusion force applied, minimum peak and 35 % MVC force) were removed from identified for EDA-1 dataset (Table 2). Similarly, unimportant features for EDA-2 (minimum peak and occlusion pressure) were eliminated. Lastly, insignificant features uncovered for EDA-3 (minimum peak and occlusion pressure) were eradicated. Interestingly, the statistical p – value ranking for EDA-1 reveal that most participants didn't consider actual experimental parameters/features eliciting pain (muscle tensile stretch, occlusion force, and 35 % MVC force) as a factor to determine pain intensity reported via VAS after the experiment (Figure 7a). Thereby buttressing the claims of inaccuracy in readings after experiments [19, 57]. Whereas Figure 7b – c indicates that some of these features were relevant to the in-session scores reported by participants reported via PSM during the experiment. The extent of this disparity on objective pain prediction and trained machine learning performance is discussed next.

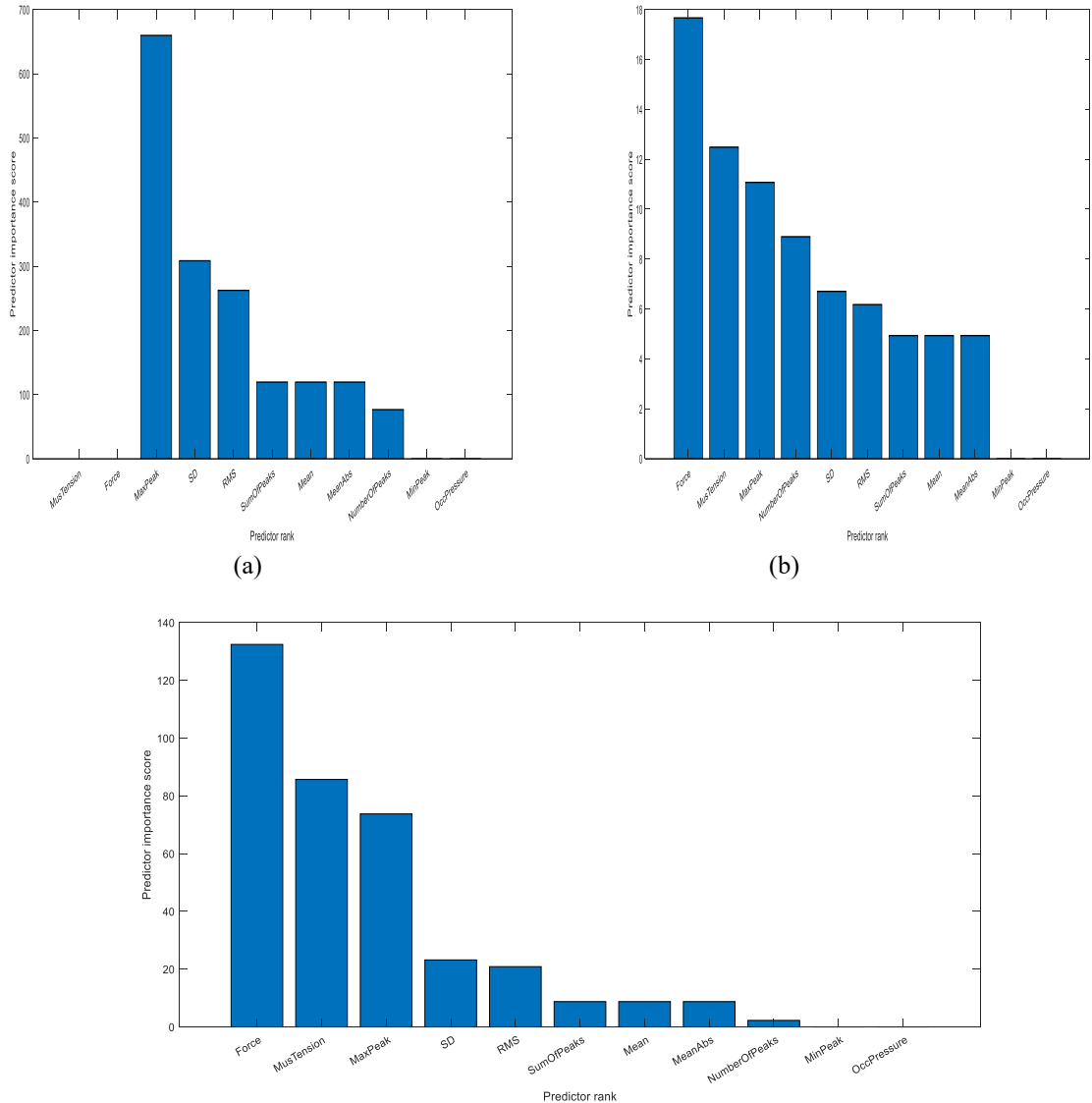

Figure 7. Filter Results for Feature Selection. (a) Feature Ranking Importance for EDA-1 Datasets (b) Feature Ranking Importance for EDA-2 Datasets (c) Feature Ranking Importance for EDA-3 Datasets

## 2.5. Machine learning Modeling

Supervised machine learning algorithms were developed for characterizing skeletal muscle pain intensity of participants based on the eliminated irrelevant objective EDA features identified with respect to each of the 3 ground truths discussed in section 2.4.2. Typically, algorithms in this category utilize input features along with ground truth to iteratively tune model parameters until a best-fit model is obtained [16]. Random forest and neural networks were selected to predict EDA features into 3 pain intensity classes (low, medium, and high) in WEKA environment. However, before this step was carried out, a categorical encoding script in MATLAB was utilized for translating normalized continuous in-session PSM Mean, PSM Mode ground truth scores and post-session VAS ground truth score into label categories of low pain intensity (0 – 0.333333333), medium pain intensity (0.333333334 - 0.666666666), and high pain intensity (0.666666667 – 1) respectively.

### 2.5.1. Neural Network

The multi-layer perceptron (MLP) is a feedforward learning algorithm that has an input layer, one or more hidden layers and an output layer. Notably, two stages called the forward and backward propagation are involved in the process of training an MLP. In the former, vectors of inputs or features are presented to the input layer for complex abstraction and learning by the hidden layer. Thereafter, the hidden layers pass forward the respective activations to the next or output layer. Conversely, depending on the output of the network, a backward propagation is conducted to update weights of the network [53]. In the present study, MLP was trained in WEKA on each of the 3 different EDA – $i$ datasets using a varying number of top (3, 4, and 7) input feature vector length, variable number of hidden layer nodes of 10, 50, 100 and 3 output classes. Sigmoid activation was used as the transfer function for the hidden and output layers.

### 2.5.2. Random Forest

Just as the name "Forest" sounds, random forest (RF) is created from combining a number of decision trees [54]. The model is generated by taking a substantial number of separate samples of data and growing "trees" out of them. Afterwards, if a new set of features describing the level of pain of an individual is presented to the forest, all the trees vote on what they feel the output should be and the final prediction is selected by the majority vote [55]. In machine learning, RF is acclaimed for prevention of over-fitting, quick processing of high dimensional data and relatively good accuracy even when datasets aren't complete [54]. Similarly, RF was trained in WEKA on each of the 3 different EDA – $i$ datasets using a varying number of top (3, 4, and 7) input feature vector length, varying bag percentage and batch size, and unlimited number of tree depth.

### 2.5.3. Performance Evaluation

ML performance evaluation in this work was done in two folds based on recommendations that traditional metrics should not be singly utilized in assessing imbalanced dataset [56]. First, WEKA's weighted average of true positive rate (TPR) or correctly classified instances was used to prune out MLP and RF model configurations that yielded low classification performance. Thereafter, the macro-averaged geometric mean score was computed from the confusion matrix of the RF and MLP configurations with seemingly good TPR performance [18, 56].

$$\text{Macro-averaged geometric mean score} = \frac{\sum_{i}^{N_{class}} \sqrt{sensitivity_i \times specificity_i}}{N_{class}} \quad (3)$$

where,
$N_{class}$ = classes representing pain
$i$ = class index.

## 3. Results

In this section, results of ML models trained with the stratified 10-fold cross validation are reported. Table 3 presents the weighted TPR accuracy metric of all MLP networks trained on each of the 3 different EDA – $i$ datasets using a varying number of top 3, 4, and 7 input feature vector length, variable number of hidden layer nodes of 10, 50, 100 and 3 output classes. Using Table 3, MLP configurations with EDA – 2 datasets + top 3 Features + 50 nodes, EDA – 2 datasets + top 4 Features + 50 nodes, EDA – 3 datasets + top 7 Features + 100 nodes, and EDA – 1 dataset

+ top 3 features + 10 nodes were selected for major evaluation using Equation 3 and their respective confusion matrix in Table 4. Results obtained using Equation 3 indicated that the selected MLP network achieved a performance score of 75.9 % and 73.1 % for the EDA – 2 datasets with 3 or 4 features + 50 nodes respectively. Additionally, a performance score of 73.3% was attained for EDA – 3 datasets with 7 features +100 nodes. Also, a 70.3 % performance score was gotten for EDA – 1 dataset with 3 features + 10 nodes (See Table 7).

Table 3. Weighted TPR Results of using Different Ground Truth, Hidden Nodes, and Feature Set on All Trained MLP Configurations.

| Dataset Type (Ground Truth Type) | Hidden nodes | Feature Vectors | | |
|---|---|---|---|---|
| | | 3 | 4 | 7 |
| EDA – 1 (VAS) | 10 | 66.06 % | 64.80 % | 64.07 % |
| | 50 | 64.44 % | 63.53 % | 63.89 % |
| | 100 | 62.27 % | 60.83 % | 64.62 % |
| EDA – 2 (PSM Mean) | 10 | 63.89 % | 64.44 % | 62.63 % |
| | 50 | 66.06 % | 66.06 % | 64.98 % |
| | 100 | 64.98 % | 64.62 % | 65.52 % |
| EDA – 3 (PSM Mode) | 10 | 59.56 % | 64.98 % | 60.83 % |
| | 50 | 62.99 % | 62.45 % | 64.98 % |
| | 100 | 62.45 % | 63.17 % | 65.88 % |

Table 4. Confusion Matrix utilized for deriving the Macro-Average Geometric Mean Scores of selected MLP models

| | | EDA – 2 (3 Features + 50 nodes) | | | EDA – 2 (4 Features +50 nodes) | | | EDA – 3 (7 Features + 100 nodes) | | | EDA – 1 (3 Features + 10 nodes) | | |
|---|---|---|---|---|---|---|---|---|---|---|---|---|---|
| | | a | b | c | a | b | c | a | b | c | a | b | c |
| Actual | a | 148 | 47 | 38 | 145 | 56 | 32 | 155 | 47 | 32 | 68 | 19 | 26 |
| Class | b | 57 | 77 | 20 | 56 | 79 | 19 | 52 | 84 | 16 | 27 | 72 | 49 |
| | c | 18 | 8 | 141 | 18 | 7 | 142 | 26 | 16 | 126 | 18 | 49 | 226 |

a = Low pain, b = Medium Pain, c = High Pain

Similarly, Table 5 shows the weighted TPR accuracy metric of all RF networks trained on each of the 3 different EDA – $i$ datasets using a varying number of top 3, 4, and 7 input feature vector length, varying bag percentage and batch size of 50, and unlimited number of tree depth. Utilizing Table 5, RF configurations with EDA – 1 dataset + 17 % bag percentage + 4 features, EDA – 2 dataset + 23 % bag percentage + 3 features and EDA – 3 dataset + 23 % bag percentage + 3 features were chosen for major assessment using Equation 3 and their respective confusion matrix in Table 6. Results obtained using Equation 3 showed that the selected RF network achieved a performance score of 78.3 % for EDA – 2 datasets + 23 % bag percentage + 3 features, 77.9 % for the EDA – 3 datasets + 23 % bag percentage + 3 features  and 74.6 % for EDA – 1 + 17 % bag percentage + 4 features (See Table 7).

Table 5. Weighted TPR Results of using Different Ground Truth, Varying Bag Percentage Size and Feature Set on All Trained RF Configurations.

| Dataset Type (Ground Truth Type) | Feature Vectors | Varying Bag Percentage for Batch Size of 50 | | |
|---|---|---|---|---|
| | | 17 | 23 | 28 |
| EDA – 2 (PSM Mean) | 3 | 70.21 % | 71.11 % | N/A |
| | 4 | N/A | 68.41 % | N/A |
| | 7 | N/A | 65.16 % | N/A |
| EDA – 3 (PSM Mode) | 3 | 70.21 % | 70.03 % | 70.93 % |
| | 4 | N/A | 66.78 % | N/A |
| | 7 | N/A | 66.06 % | N/A |
| EDA – 1 (VAS) | 3 | 72.38 % | 71.29 % | 71.66 % |
| | 4 | 73.28 % | N/A | N/A |
| | 7 | N/A | 71.48 % | N/A |

Table 6. Confusion Matrix utilized for deriving the Macro-Average Geometric Mean Scores of selected RF models

| | | EDA – 2 (3 Features + 23 % Bag) | | | EDA – 3 (3 Features + 28 % Bag ) | | | EDA – 1 (4 Features + 17 % Bag ) | | |
|---|---|---|---|---|---|---|---|---|---|---|
| | | a | b | c | a | b | c | a | b | c |
| Actual Class | a | 152 | 50 | 31 | 152 | 50 | 32 | 71 | 15 | 27 |
| | b | 32 | 99 | 23 | 34 | 100 | 18 | 7 | 83 | 58 |
| | c | 14 | 10 | 143 | 17 | 10 | 141 | 13 | 28 | 252 |

a = Low pain, b = Medium Pain, c = High Pain

Table 7. Macro Averaged Geometric Mean Performance of ML models trained on PSM In-Session Scores and VAS Post-Session Scores

| Dataset Type (Ground Truth Type) | Feature Vectors | Random Forest 50 % Batch Size | | MLP varying number of nodes | | |
|---|---|---|---|---|---|---|
| | | 17 | 23 | 10 | 50 | 100 |
| EDA – 2 (PSM Mean) | 3 | N/A | 78.3 % | N/A | 75.9 % | N/A |
| | 4 | N/A | N/A | N/A | 73.1 % | N/A |
| | 7 | N/A | N/A | N/A | N/A | N/A |
| | 3 | N/A | 77.9 % | N/A | N/A | N/A |
| EDA – 3 (PSM Mode) | 4 | N/A | N/A | N/A | N/A | N/A |
| | 7 | N/A | N/A | N/A | N/A | 73.3 % |
| EDA – 1 (VAS) | 3 | N/A | N/A | 70.3% | N/A | N/A |
| | 4 | 74.6 % | N/A | N/A | N/A | N/A |
| | 7 | N/A | N/A | N/A | N/A | N/A |

## 4. Discussion

Prior studies have purported the idea of a more reliable pain ground truth would be advantageous, as it would address the issue of scanty scores, label noise and improved learned models for each participant's objective data [13, 14, 17]. As a result, this study set out with the aim of judging if machine learning algorithms trained for objective pain characterization would yield better performance when continuous in-session ground truth scores are utilized over post-session ground truth scores. The study further evaluated whether the maximum (PSM-Mode) or integral (PSM-Mean) of in-session continuous ground truth scores, extracted alongside objective EDA features over a non-overlapping window length of 10, outperformed single VAS self-report scores. This evaluation was crucial because pain does not suddenly cease after a stimulus but diminishes gradually [19]. Additionally, relying solely on participants' single self-report scores could negatively impact ML performance compared to ground truth scores that encompass the cumulative pain intensity experienced or the maximum pain intensity that arose in phases during an experiment [57]. Achieving these objectives required the design of a novel ML-centric, continuous in-session ground truth acquisition device called PSM. This device was interfaced with a data-acquisition device (PainGad) that simultaneously acquires and stores corresponding PSM (0 – 10) and EDA signals into a platform (PainPlat) at frequency of 2 Hz.

Results obtained from several experiments for trained MLP models reveal that EDA – 2 dataset + 3 features + 50 nodes and EDA – 2 dataset + 4 features + 50 nodes surpassed EDA – 1 dataset + 3 features + 10 nodes by 5.6 % and 2.8 %. In similar vein, EDA – 3 dataset + 7 features + 100 nodes bettered EDA – 1 dataset + 3 features + 10 nodes by 3.0 % (Table 7). To buttress the efficacy of these findings, Table A1 reveals further that the VAS had the lowest correctly classified instances for all classes of pain. In sum, these findings validate the theories that declare accuracies of trained ML systems are greatly affected by scores from traditional scales (VAS) and unreliable ground truth [17, 18]. Another finding attained from several experiments of RF models is that the best classifier performance of 78.3 % was obtained using EDA – 2 datasets + 3 features + 23 % bag percentage. In comparison with RF models trained with EDA – 1 datasets + 4 features + 17 % bag percentage, a 3.7% increase in performance was obtained. In like manner, it was observed that EDA – 3 datasets + 3 features + 23 % percentage outperformed EDA – 1 datasets + 4 features + 17 % bag percentage by an upward rise of 3.3 % (Table 7). Surprisingly, when the correctly classified instances of each class were examined in Table A2, EDA – 2 and EDA – 3 modalities performed better than the VAS (EDA – 1) by having a high classification accuracy for low and medium pain intensity class. On the contrary, VAS ground truth outperformed the EDA – 2 and EDA – 3 ground truths for the high pain intensity class alone. A note of caution to bear in mind at this point is that Table A2 was assessed based on the TPR metric that is susceptible to datapoint imbalance in classes [56, 58]. However, a possible explanation for this anomaly is that 80% of the participants penned down high pain on the VAS scale after the experiment leaving other classes imbalanced as complained by prior authors in [13, 14, 17, 18]. Therefore, it is tricky to conclude that the VAS bettered the PSM ground truth approach especially when the in-session PSM device captured more dataset distribution for all classes.

On this note, individual TPR was investigated for each class of the selected RF models. It was evident that the EDA – 2 ground truth provided a closely matched accuracy of 85 % when compared to the VAS accuracy of 86 % for high pain intensity prediction (see Table A2). Most importantly, approach EDA – 2 yielded better accuracy of 64 % and 65 % when compared to VAS's 62 % and 56 % for medium and low pain intensity prediction respectively. This evidence suggests that achieving robust ML performance across all classes requires a tradeoff between PSM benefits (improved performance, better dataset distribution) and VAS perks (addressing sparsity and dataset imbalance in some classes). To the best of authors knowledge, there are no studies that have considered acquisition of continuous in-session ground truth for improving ML performance. Nevertheless, the results in this work align with prior research using periodic in-session or post-session ground truth scores, showing improved ML performance over [18, 22] (i.e., pain stimulation intensity classification and multimodal objective human pain intensity estimation works). In the future, researchers may employ a dual approach to training and obtaining high accuracy in ML pain predictive system.

## 5. Conclusions

The present study makes several noteworthy contributions to improving ML performance through reliable ground truth acquisition. First, it extends our knowledge that acquiring in-session ground truth scores (EDA – 2 and EDA – 3) produces continuous subjective trend of pain (Figure 6a) that reduces label randomness that occurs due to sparsity in scales like VAS, NRS and VRS. Secondly, this work mitigates latent introduction of high variance that causes poor performance of ML models and datapoint imbalance in each class. Also, we prove that the maximum or integral of continuously acquired in-session ground truth scores betters the traditional VAS score for ML based pain predictive system. Nonetheless, several limitations to this pilot study need to be acknowledged. In comparison to [18 - 32], the entire duration of experiment is considered to be small especially if more pain level classes are desired. Additionally, the study has only examined pain arising from skeletal muscle afferents under a controlled PECO + stretch experiments. A future progression of this work would involve conducting and testing the new continuous in-session ground truth score acquisition method on other large scale and controlled pain trials for more assertive evidence that this approach results in better ML performance.

**Author Contributions:**
**B.E.F:** Conceptualization of this study, hardware and instrumentation, web application software and database storage, data processing pipelines and machine learning, manuscript writing, result interpretation; **A.H.S** Machine learning and methodology, conceptualization of this study, manuscript revision, result interpretation **Z.Z:** Machine learning, database, hardware component sourcing, manuscript revision **J.S:** IRB approval, study methodology and experiment design, interpretation of results from experiments, manuscript revision **N.N:** Study design, experiment, data interpretation, manuscript revision.

**Institutional Review Board Statement:**
The study was approved by the Institutional Review Board of University of Southern Mississippi (protocol code 22-1012 Approved: 17-November-2022).

**Informed Consent Statement:**
All participants gave an informed consent.

**Data Availability Statement:**
Data utilized in this study is available on request from the corresponding author.

**Acknowledgement:**
The authors would like to acknowledge the contributions of Atawodi Ilemona during ML experiment phase.

**Conflicts and Interest:**
The authors declare no conflict of interest.

**Appendix**
Table A1 presents detailed classifier report for MLP networks with highest weighted TPR accuracies and Table A2. detailed classifier report for RF networks with highest weighted TPR accuracies.

Table A 1. Detailed Classifier Report for MLP Networks with Highest Weighted TPR Accuracies

| Dataset (Ground Truth) Type | TP Rate | FP Rate | Precision | Recall | F-Measure | ROC Area | Class |
| --- | --- | --- | --- | --- | --- | --- | --- |
| **EDA – 2 (PSM Mean) + 4 Features + 50 nodes** | 0.62 | 0.23 | 0.66 | 0.62 | 0.64 | 0.75 | Low Pain |
| | 0.51 | 0.15 | 0.55 | 0.51 | 0.53 | 0.76 | Medium Pain |
| | 0.85 | 0.13 | 0.73 | 0.85 | 0.78 | 0.90 | High Pain |
| **EDA – 2 (PSM Mean) + 3 Features + 50 nodes** | 0.63 | 0.23 | 0.66 | 0.63 | 0.64 | 0.75 | Low Pain |
| | 0.50 | 0.13 | 0.58 | 0.50 | 0.53 | 0.79 | Medium Pain |
| | 0.84 | 0.15 | 0.70 | 0.84 | 0.77 | 0.90 | High Pain |
| **EDA – 3 (PSM Mode) + 7 Features + 100 nodes** | 0.66 | 0.244 | 0.66 | 0.66 | 0.66 | 0.76 | Low Pain |
| | 0.55 | 0.157 | 0.57 | 0.55 | 0.56 | 0.76 | Medium Pain |
| | 0.75 | 0.124 | 0.72 | 0.75 | 0.73 | 0.88 | High pain |
| **EDA – 1 (VAS) + 3 Features + 10 nodes** | 0.60 | 0.10 | 0.60 | 0.60 | 0.60 | 0.86 | Low Pain |
| | 0.48 | 0.16 | 0.51 | 0.48 | 0.50 | 0.72 | Medium Pain |
| | 0.77 | 0.28 | 0.75 | 0.77 | 0.76 | 0.83 | High Pain |

Table A 2. Detailed Classifier Report for RF Networks with Highest Weighted TPR Accuracies

| Dataset (Ground Truth) Type | TP Rate | FP Rate | Precision | Recall | F-Measure | ROC Area | Class |
| --- | --- | --- | --- | --- | --- | --- | --- |
| **EDA – 2 (PSM Mean) + 3 Features + 23 % Bag** | 0.65 | 0.14 | 0.76 | 0.65 | 0.70 | 0.81 | Low Pain |
| | 0.64 | 0.15 | 0.62 | 0.64 | 0.63 | 0.81 | Medium Pain |
| | 0.85 | 0.14 | 0.72 | 0.85 | 0.78 | 0.92 | High Pain |
| **EDA – 3 (PSM Mode) + 3 Features + 28 % Bag** | 0.65 | 0.15 | 0.74 | 0.65 | 0.69 | 0.80 | Low Pain |
| | 0.65 | 0.14 | 0.62 | 0.65 | 0.64 | 0.82 | Medium Pain |
| | 0.83 | 0.13 | 0.73 | 0.83 | 0.78 | 0.92 | High pain |
| **EDA – 1 (VAS) + 4 Features + 17 % Bag** | 0.62 | 0.04 | 0.78 | 0.62 | 0.69 | 0.92 | Low Pain |
| | 0.56 | 0.10 | 0.65 | 0.56 | 0.60 | 0.85 | Medium Pain |
| | 0.86 | 0.32 | 0.74 | 0.86 | 0.80 | 0.88 | High Pain |